\documentclass[conference]{IEEEtran}
\usepackage{times}

\usepackage[numbers]{natbib}
\usepackage{multicol}
\usepackage[bookmarks=true]{hyperref}

\usepackage{amsmath,amssymb,amsfonts}
\usepackage{algorithmic}
\usepackage{graphicx}
\usepackage{textcomp}
\usepackage[table]{xcolor}
\usepackage{comment}
\def\BibTeX{{\rm B\kern-.05em{\sc i\kern-.025em b}\kern-.08em
    T\kern-.1667em\lower.7ex\hbox{E}\kern-.125emX}}
\usepackage{hyperref}
\usepackage{cleveref}[2012/02/15]

\begin{document}

\title{Sense and Sensibility: What makes a social robot convincing to high-school students?}
\author{
    \IEEEauthorblockN{Pablo González-Oliveras\IEEEauthorrefmark{1}, Olov Engwall\IEEEauthorrefmark{1}, Ali Reza Majlesi\IEEEauthorrefmark{2}} \\
    \IEEEauthorblockA{\IEEEauthorrefmark{1}Dept. of Intelligent Systems, KTH Royal Institute of Technology, Stockholm, Sweden} \\
    \IEEEauthorblockA{\IEEEauthorrefmark{2}Dept. of Education, Stockholm University, Stockholm, Sweden} \\
    pablool@kth.se, engwall@kth.se, ali.reza.majlesi@edu.su.se
}

\maketitle

\begin{abstract}
This study with 40 high-school students demonstrates the high influence of a social educational robot on students' decision-making for a set of eight true-false questions on electric circuits, for which the theory had been covered in the students' courses.
The robot argued for the correct answer on six questions and the wrong on two, and 75\% of the students were persuaded by the robot to perform beyond their expected capacity, positively when the robot was correct and negatively when it was wrong.
Students with more experience of using large language models were even more likely to be influenced by the robot’s stance -- in particular for the two easiest questions on which the robot was wrong -- suggesting that familiarity with AI can increase susceptibility to misinformation by AI. 

We further examined how three different levels of portrayed robot certainty, displayed using semantics, prosody and facial signals, affected how the students aligned with the robot's answer on specific questions and how convincing they perceived the robot to be on these questions. The students aligned with the robot’s answers in 94.4\% of the cases when the robot was portrayed as Certain, 82.6\% when it was Neutral and 71.4\% when it was Uncertain. The alignment was thus high for all conditions, highlighting students’ general susceptibility to accept the robot’s stance, but alignment in the Uncertain condition was significantly lower than in the Certain. Post-test questionnaire answers further show that students found the robot most convincing when it was portrayed as Certain. These findings highlight the need for educational robots to adjust their display of certainty based on the reliability of the information they convey, to promote students' critical thinking and reduce undue influence.


\end{abstract}





\IEEEpeerreviewmaketitle

\section{Introduction}\label{sec:Introduction}


\noindent Educational robots are becoming more common and they have significant potential in, e.g., STEM (science, technology, engineering and mathematics) education \cite{Ouyang2024, Zhong2018, Donnerman2022}, offering students realistic and natural interactions, not the least by employing Large Language Models (LLMs), as demonstrated in several recent studies \cite{mishra2023real,Zhang2023,Ye2023}.  
However, it is also well-known that while the LLMs’ linguistic proficiency is often astonishing, their factual ''knowledge'' in STEM subjects is flawed, and incorrect statements occur frequently \cite{Kuchemann2023,Steinert2023}.
Since robots can exert high informational social influence \cite{liu2022systematic,hertz2016influence,hertz2018under,Salomons2018humans,salomons2021minority} and students will align with the robot's views to large extents \cite{Kamelabad2024}, the positive as well as negative effects of learning with a social robot need to be considered: 
\textit{Students} need to use critical thinking to decide if they should accept the robot’s propositions \cite{vanDerWerff2019}. \textit{Educators} need to understand which students are more at risk of being misled by a robot presenting incorrect STEM facts, to provide in-time support. 
\textit{Developers} need to find ways to signal how certain the robot is about the presented facts to avoid overtrust \cite{Robinette2016}.

In a prior exploratory study (currently submitted), we found that \textit{Persuasion}, i.e., robot arguments for an incorrect solution to a maths problem, and \textit{Prejudice}, i.e., students' positive attitudes towards robots and more experience of using LLMs, influenced a large majority of students to conform with the robot's incorrect solution. 
On the other hand, taking \textit{Pride} in being a strong maths student increased resistance against incorrect arguments.

The present study follows up on these findings, by systematically investigating \textit{Sense}, i.e., the students' \textit{''reliable ability to judge and decide with soundness, prudence, and intelligence''} [Merriam-Webster] if the robot's arguments are correct or incorrect; and \textit{Sensibility}, i.e., the students' \textit{''awareness of and responsiveness toward [\dots] emotion in another''} [Merriam-Webster] regarding their responses to the robot's arguments depending on the robot's multimodal display of certainty presented using semantics, prosody and facial signals.

\subsection{Study objectives}
\label{sec:INTRO-RQ}
\noindent To guide this study, we posed the following research questions:
\begin{itemize}
    \item \textbf{RQ1}: To what extent are high-school students influenced by a social robot’s correct or incorrect arguments during a series of true/false questions?
    \item \textbf{RQ2}: How does the robot’s expressed certainty (uncertain, neutral, certain) affect the likelihood that students align with its answers?
    \item \textbf{RQ3}: Does the robot’s behavior on preceding questions (being right or wrong; certain or uncertain) influence student alignment in later interactions?
    \item \textbf{RQ4}: Do personal characteristics—such as extroversion, self-perception as a learner, or prior experience with AI—affect the likelihood of students aligning with the robot?
\end{itemize}

Our expectations, based on our previous exploratory study, are that 
(\textbf{H1}) a majority of students will align with the robot's answer even when it is incorrect; 
(\textbf{H2}) students will be more prone to follow the robot when it is portrayed as being certain.
(\textbf{H3}) robot argumentation on preceding questions will, to some extent, influence students to follow the robot or not;
(\textbf{H4})
students with greater AI experience will align more frequently
with the robot’s answers, even when they are incorrect.



\section{Background \& Related work}\label{sec:Related}


\noindent Like human actors, social robots impact their interaction partners through informational and normative social influence \cite{Zonca2021, beckner2016participants, brandstetter2014peer}.
These processes can lead to conformity or persuasion, which differ in how the target perceives the source's intent. In conformity, individuals adjust their stance after being exposed to others' opinions, without perceiving an active attempt to change their minds \cite{deutsch1955}. In persuasion, the target typically perceives an intentional effort to influence their stance, which can involve explicit argumentation and reasoning, appeals to emotion, or credibility cues \cite{Cialdini2004, Kelman1958}.

This study focuses on how informational trust (RQ1) and the robot's persuasive certainty cues (RQ2) affect students' willingness to align with the robot's answers, while also exploring whether prior interactions influence subsequent decisions (RQ3) and whether individual traits like AI experience moderate these effects (RQ4). The following subsections elaborate these constructs.

\subsection{Informational Trust in HRI}
\label{sec:rw-infotrust}
\noindent Informational trust plays a key role in informational social influence, as individuals rely on perceived reliable sources when making decisions under uncertainty \cite{HANCOCK}.
Cognitive dissonance theory suggests that individuals weigh new information against existing beliefs, with higher uncertainty leading to increased reliance on external sources such as robots \cite{lee2021students}. 
This dynamic is especially relevant in educational HRI, where students often perceive robots as authoritative sources of knowledge \cite{chidambaram2012designing}.
As a result, overtrust can lead students to adopt information even when the robot is clearly wrong \cite{braaten2009trust, hare2007credibility}.  

A meta-analysis \cite{HANCOCK} found that factors such as reliability, false alarm rates, and failure rates significantly influence trust development in HRI. Robots that adapt to a student’s learning pace, tailor explanations, and acknowledge user input are perceived as more trustworthy \cite{okamura2020adaptive, park2018influence}. 
Peer-like, anthropomorphic robots, such as Furhat used in this study, further enhance trust through human-like gestures and  responsiveness \cite{BREAZEAL, RAE}, particularly when their information is consistent and accurate \cite{nowak2023subjective}.

In this study, students' ability to critically assess the robot’s answers is shaped by their prior knowledge of the topic. Their evaluation of the robot’s correctness, along with their perception of its confidence, determines the level of informational trust they place in it, ultimately influencing how susceptible they are to its argument-based persuasion.


\subsection{Persuasion in HRI}
\label{sec:rw-persuasion}
\noindent 
Argument-based persuasion is explained by the Elaboration-Likelihood Model, which describes how persuasion can influence attitudes via two distinct routes: the central route, relying on detailed reasoning, and the peripheral route, dependent on superficial cues like authority or likability \cite{petty1986elaboration}.
A 2022 systematic review confirmed that while persuasion has been extensively explored in HRI \cite{liu2022systematic},  most studies focus on peripheral strategies such as compliance, assertiveness, and emotional cues \cite{Paradeda2020, Alam2021}, with limited attention to logical argumentation. 
Among 54 identified studies, only one partially addressed structured argumentation \cite{Saunderson2019} and this trend has persisted in recent years \cite{Saunderson2021, saunderson2020investigating}.
Studies on persuasion by educational robots are even scarcer. 
A systematic review of 89 studies found only two studies addressing real-world educational tasks \cite{BELPAEME-REVIEW}. These studies demonstrated positive effects on compliance \cite{bainbridge2011benefits} and conformity \cite{kennedy2015comparing}, but their designs lacked generalizability beyond experimental academical contexts.
Moreover, the review reaffirmed the heavy focus on the peripheral route and revealed that approximately two-thirds of the reviewed studies focused on affective rather than cognitive outcomes. 

\subsection{Challenges to Students' Critical Thinking in HRI
}
Learning gains in educational HRI are often measured via pre- and post-tests \cite{BELPAEME-REVIEW}, neglecting  students’ academic uncertainty and vulnerability of their existing knowledge. 
Further, few studies have explored the interaction between robot confidence cues and learner certainty.
This highlights a gap in studies addressing how robots and students interact with uncertain or conflicting prior knowledge.


Recent studies have shown the potential of tailored persuasive strategies, such as personalized storytelling to align with users’ traits \cite{Paradeda2020}, or adjusting assertiveness to suit social contexts \cite{paradeda2020importance}. However, they also highlight risks like persuasive backfiring, where ineffective emotional appeals or excessive assertiveness can undermine trust in robots \cite{Alam2021}. 
These risks have become more prominent with the recent development of LLM-driven educational robots that have transformed interactions with learners.
They improve feedback and guidance during problem-solving processes \cite{Zhang2023, Stella2023, Steinert2023}, but since these robots tend to increase students’ trust in their outputs \cite{Ye2023} and the factual correctness of LLMs in STEM contexts has been questioned \cite{Kuchemann2023, Avila2024}, there is a need for students to critically evaluate AI-generated information, which they often fail to do \cite{Krupp2024}. 
Techniques such as adaptive feedback, enhanced multimodal communication, and transparency in decision-making can 
improve improve students' assessment of their interaction with potentially unreliable robots \cite{Oksanen2020, Winkle2019, salomons2021minority, Kamelabad2024}.

This study addresses research gaps by investigating how variations in a robot’s displayed confidence influence the outcome of the robot's argument-based persuasion.
Based on these constructs, we hypothesize that (H1) informational trust will lead students to align even when the robot is incorrect, (H2) robot certainty will enhance alignment through persuasive cues, (H3) prior robot behavior might affect future alignment via informational updating, and (H4) students' AI familiarity and personality traits will moderate these effects.

\section{Methodology}\label{sec:Methods}
\begin{figure}[t]
    \centering
    \includegraphics[width=1\linewidth]{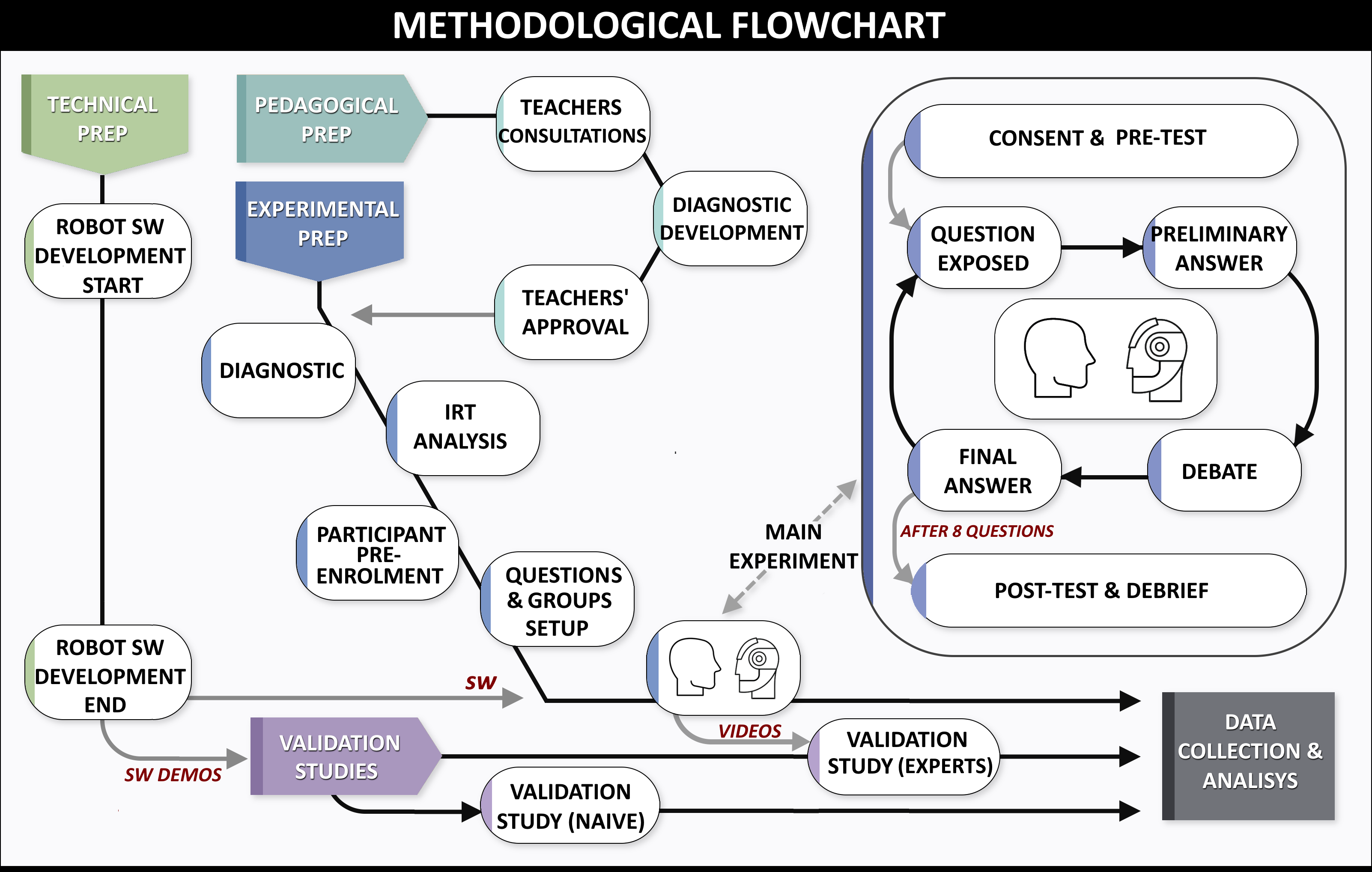}  
    \caption{Methodological flowchart outlining study phases: diagnostic test, robot interaction, and pre/post-questionnaires.}
    \label{fig:flowchart}
\end{figure}

\begin{figure}[bt]
    \centering
    \includegraphics[width=1\linewidth]{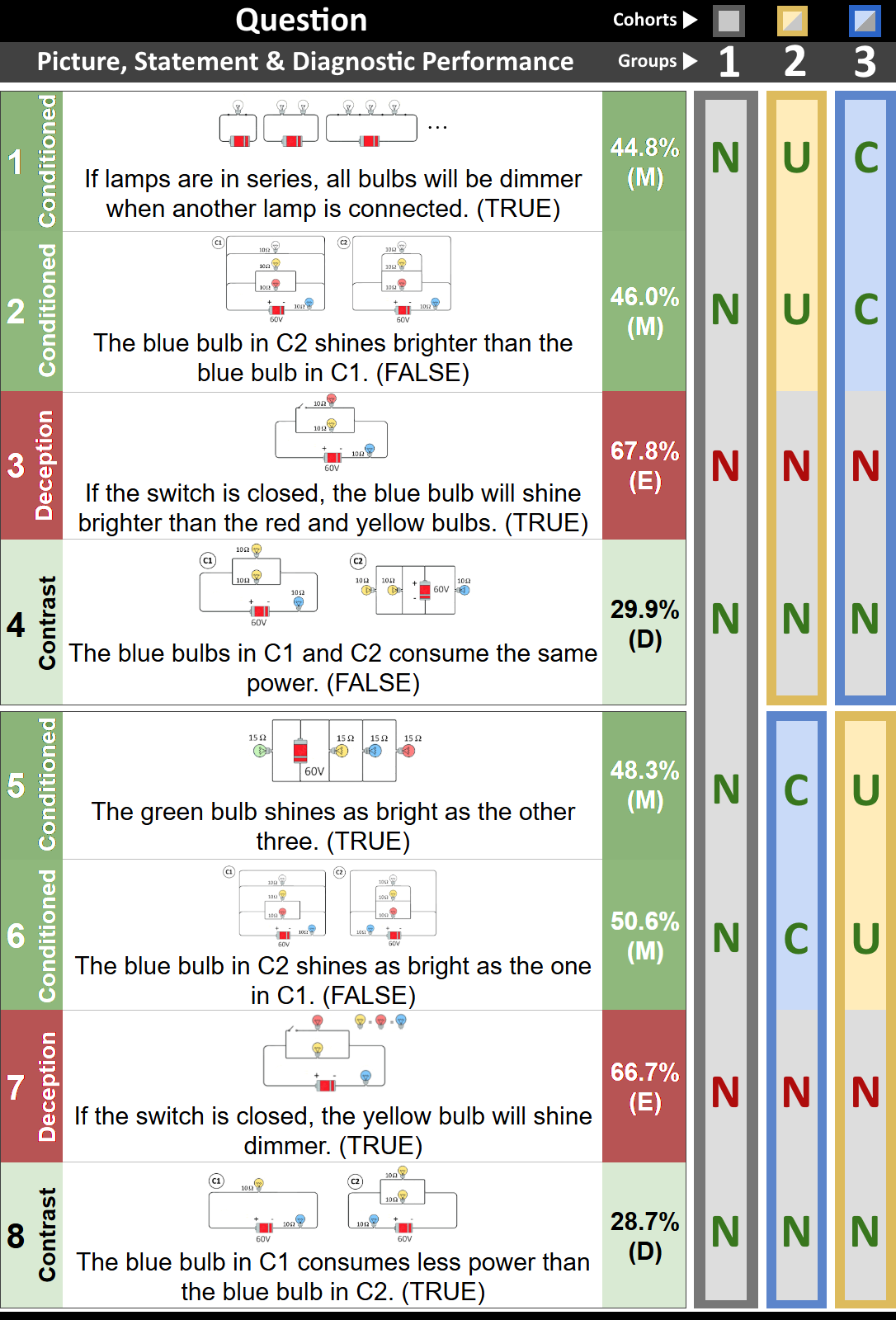}
    \caption{The eight electric circuit statements with ratio of correct  diagnostic answers (E=Easy, M=Medium, D=Difficult). Robot certainty levels indicated as Certain (C, pale blue background), Neutral (N, light gray background)  and Uncertain (U, soft yellow background), along with correctness (\textcolor{teal}{correct} or \textcolor{red}{wrong}). Cohort frames are coloured dark gray (N), yellow (U), and blue (C).}
    \label{fig:electric-problems}
\end{figure}

\noindent We designed an experiment in which Swedish secondary school students interacted one-on-one with a fully autonomous educational robot, discussing eight electric circuits to determine whether statements about each were true or false (see Fig.~\ref{fig:electric-problems}).
The overall procedure is summarized in the methodological flowchart (Fig.~\ref{fig:flowchart}), which outlines the sequence from diagnostic testing to robot interaction and post-session assessments.

The recruitment of students and the questions were planned together with two secondary school teachers to ensure that topics corresponded to material that had been covered in the students' classes.
One month before the experiment the teachers distributed a diagnostic test with 19 three-choice questions about electric circuits in their classes, respectively in grade 10, 11 and 12 of a practically oriented electrical engineering program, and grade 11 and 12 of a theoretical natural sciences program. The students were unaware that the diagnostic test was linked to the upcoming experiment. 

The diagnostic test had two main objectives: 
1) selecting the eight questions for the experiment and 
2) assessing student subject knowledge, to balance group distribution and evaluate their performance in the main experiment.
An Item Response Theory analysis of the diagnostic test answers (\textbf{DA}) was then conducted to assess both student ability and question difficulty, as input to the experimental design.


\subsection{Item Response Theory}
\label{sec:INTRO-IRT}
\noindent The item response theory (\textbf{IRT})\footnote{https://www.publichealth.columbia.edu/research/population-health-methods/item-response-theory}
refers to a family of mathematical models that attempt to explain the relationship between latent traits (unobservable characteristic) 
and their observed outcomes.
We opted for the 3-Parameter Logistic (\textbf{3PL}) model, 
which, in addition to difficulty ($b$) and discrimination ($a$) includes a guessing parameter ($c\geq0.25$ to prevent overfitting), making it particularly appropriate for three-option multiple-choice tests to account for guessing by low-ability students. 
The following IRT assumptions were applied:
\textit{monotonicity} (the probability of a correct response increases with increased knowledge of electric circuits); \textit{unidimensionality} (the dominant latent trait, ability, is the driving force for the observed DA); \textit{local independence} (separate DA are mutually independent given a certain level of ability) and \textit{invariance} (parameters can be estimated from DA for any sub-group meeting the conditions).

The analysis was implemented in a two-stage process in a Python-driven Jupyter Notes environment. The first stage involved iteratively estimating the Item Characteristic Curves (\textbf{ICCs}) for each question. This provided metrics for item difficulty ($a$), discrimination ($b$), and guessing ($c$). Initial parameters and bounds were set and refined through multiple runs of the algorithm to achieve optimal values. 
Akaike Information Criterion (AIC) and Bayesian Information Criterion (BIC) were used as indicators of model fit, guiding the iterative parameter adjustment to identify the best model configuration.

In the second stage, the calibrated ICC parameters 
were used to estimate each student’s latent ability and the probability of a correct response for both DA (for verification) and experiment (for analysis).
In the 3PL model, the probability that a student with ability $\theta$ will correctly answer a specific question is
$p(\theta) = c + \frac{1 - c}{1 + e^{-a(\theta - b)}}$.
This probability can be used to predict the students' base performance in the experiment and thus assess the robot's influence.
DA was lacking for four students and their ability and probable correctness per question was instead estimated through a second IRT analysis using the 40 students' preliminary answers (\textbf{PA}) in the interaction with the robot.
The validity of these metrics was confirmed through a correlation analysis (r=0.43, p=.009) between DA and PA.

\subsection{Robot conditions}
\noindent
As outlined in the research questions, this study investigated how robot correctness and certainty influence participants’ responses, using three levels of portrayed robot certainty as experimental conditions. This section describes these robot conditions and their validation.

The robot was portrayed as having three levels of certainty, Uncertain \textbf{\emph{U}}, Neutral \textbf{\emph{N}}, and Certain \textbf{\emph{C}}.
These portrayals were achieved through semantic, prosodic, and facial cues: adjustments of arguments (Sec~\ref{sec:robot-arguments}), speech rate \cite{Kirkland2023} (-10\% for \(U\) and +10\% for \(C\) relative to \(N\)), insertion of pauses \cite{Lameris2023} for \(U\), and facial expressions \cite{Vincze2016}. 
In the \(U\) condition, the robot used filled and silent pauses, subtle smiles, slow gaze shifts, head tilts, half-closed eyes, and pursed lips. In the \(C\) condition, it exhibited open smiles, wider eyes, raised eyebrows, direct gaze, and slow nodding. 

\begin{figure}[b]
    \centering
    \includegraphics[width=1\linewidth]{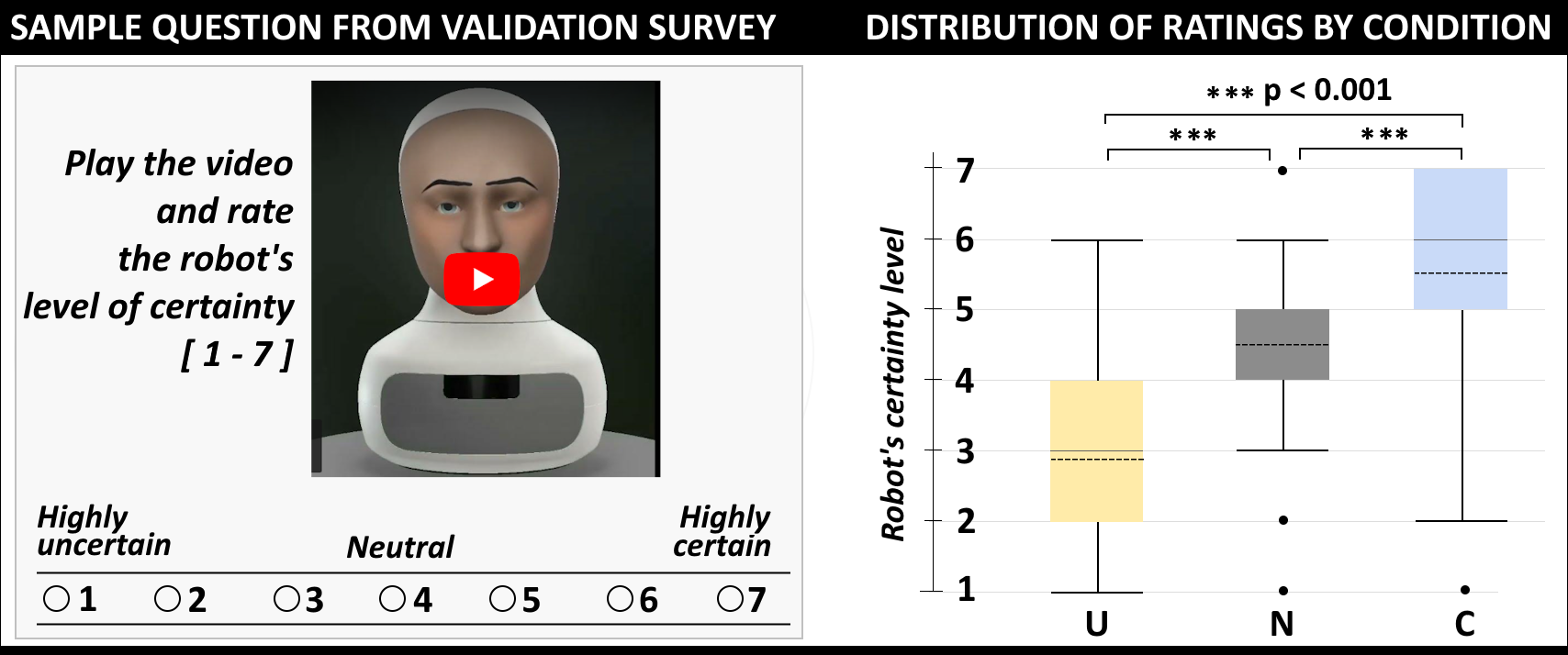}
    \caption{Left: Sample question from the validation survey interface. Right: Distribution of certainty ratings for the three robot conditions (\(U\), \(N\), \(C\)), with significant differences between conditions.}
    \label{fig:validation-boxplot}
\end{figure}

An online \textit{validation survey} was conducted to ensure that the robot’s \(U\), \(N\) and \(C\) portrayals were perceived as intended. 
Fifteen short video clips (less than 8~s each) were created, representing five distinct contexts (\textbf{\emph{c}1}--\textbf{\emph{c}5}) combined with the three certainty levels (\(U\), \(N\), \(C\)), showing the robot saying the utterances in Table~\ref{tab:validation-scenarios} in randomized order. 
The contexts consisted of the robot disclosing its answer before knowing the student's response (\(c1\)), disclosing while disagreeing (\(c2\)) or agreeing (\(c3\)) with the student, reminding the student of its position (\(c4\)), and presenting an argument to support its answer (\(c5\)). The utterances, which also occurred during the main experiment, were delivered by the robot looking directly at the camera against a black backdrop, as shown in Fig.~\ref{fig:validation-boxplot}.

\begin{table*}[t]
    \centering
    \caption{Experiment-used robot utterances chosen for the validation survey, covering five distinct situations (\(c1\)--\(c5\)) for conditions N (''Neutral''), U (''Uncertain'') and C (''Certain''). ''\dots'' denotes pauses.}
    \label{tab:validation-scenarios}
    \begin{tabular}{c}
        \includegraphics[width=\textwidth]{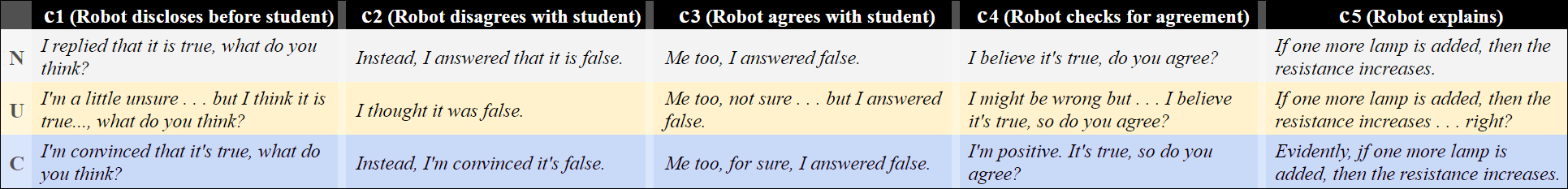}
    \end{tabular}
\end{table*}


Invitations were sent to 323 MSc, 166 BSc and 57 PhD students at a technical university.
The survey began by informing about the purpose and procedures and the participants then self-assessed their proficiency in Swedish on a 6-point Likert scale. 
For the validation, we filtered the 125 responses to only include subjects reporting medium to high proficiency (the top three levels).
This resulted in 76 participants, of which 15 were excluded for incomplete responses and 2 for failing to watch all videos.
59 participants thus rated the robot’s certainty on a 7-point Likert scale, where 1 indicated 'Highly Uncertain,' 4 'Neutral,' and 7 'Highly Certain' (see Fig.~\ref{fig:validation-boxplot}, left), resulting in 885 ratings (295 per condition).

Statistical analysis confirmed that the \(U\) condition (\(\mu = 2.86, \sigma = 1.30\)) was perceived as less certain than the \(N\) (\(\mu = 4.45, \sigma = 1.39\)) and \(C\) conditions (\(\mu = 5.50, \sigma = 1.36\)). 
A one-way ANOVA revealed significant differences between conditions (\(F = 285.31, p < 0.001\)), and post-hoc Tukey tests confirmed that all pairwise differences were statistically significant (\(p < 0.001\)), as shown in Fig.~\ref{fig:validation-boxplot} (right). Cronbach’s \(\alpha = 0.74\) indicated acceptable internal consistency, supporting the reliability of participants' assessments.
It can be noted that, in qualitative terms, the \(U\) condition was perceived as only slightly uncertain, the \(N\) condition as slightly more certain than neutral and the \(N\) condition as moderately certain.

\subsection{Subjects, Groups \& Cohorts}
\noindent 
A modified mixed factorial design was used \cite{Montgomery2022}, with robot correctness (within subjects, RQ1) and robot certainty (between and within subjects, RQ2--3) as factors. The robot’s certainty levels were varied across three groups of participants during the experiment. Group 1 (baseline) always experienced the robot with neutral certainty (\(N\) condition), while Groups 2 and 3 alternately experienced robot uncertainty (\(U\)) or certainty (\(C\)) after giving their preliminary answer.

47 subjects registered for the experiment and they were distributed between groups by means of stratified assignment, using the characteristics educational program, gender, grade level and ability as criteria, categorised in canonical combinations to assign students to groups.
40 subjects (31 male, 8 female, 1 non-binary, average age 17.7$\pm$0.86 years) of the 47 showed up.
28 were from the practical Electrical Engineering (\textbf{E}) and 12 from the theoretical Natural Sciences (\textbf{Na}) program with 9, 15 and 16 students respectively from grades 10--12.
Due to the no-shows, the distribution became unbalanced:\\
\textit{Group 1:} n=14 (11 Male, 3 Female; 10 E, 4 Na), DA performance (completed by 13 subjects): $m_{DA:13}$=7.4$\pm$1.9. \\
\textit{Group 2:} n=11 (8M, 3F; 7E, 4Na), $m_{DA:9}$=8.4$\pm$0.97.  \\
\textit{Group 3:} n=15 (12M, 2F, 1N/A; 11E, 4Na), $m_{DA:14}$=8.9$\pm$1.7. \\
As the diagnostic score was similar for Groups 2 and 3 but substantially lower for Group 1, a data analysis is required controlling for abilities calculated through IRT as a covariate.

Three student cohorts were created, as illustrated in Fig.~\ref{fig:electric-problems}.
\textit{\textbf{Cohort N}} corresponds to Group 1, exposed to \textit{Neutral} condition throughout.
\textit{\textbf{Cohort U}} consists of Group 2 for questions Q1–4 and Group 3 for Q5–8, interacting with an \textit{Uncertain} robot.
Conversely, \textit{\textbf{Cohort C}} includes Group 3 for questions Q1–4 and Group 2 for Q5–8, interacting with a \textit{Certain} robot.
Alternating Groups 2 and 3 between \textit{Cohorts U} and \textit{C} served two purposes: first, to create a more natural interaction, preventing individual students from experiencing a robot that was \textit{always} certain or always uncertain; and second, to enable the investigation of how robot certainty levels on preceding questions influenced students' subsequent responses (RQ3).



\begin{table*}[t]
    \centering
    \caption{Examples of robot arguments for conditions N (''Neutral''), U (''Uncertain'') and C (''Certain''). ''\dots'' denotes pauses.}
    \label{tab:arguments}
    \begin{tabular}{c}
        \includegraphics[width=\textwidth]{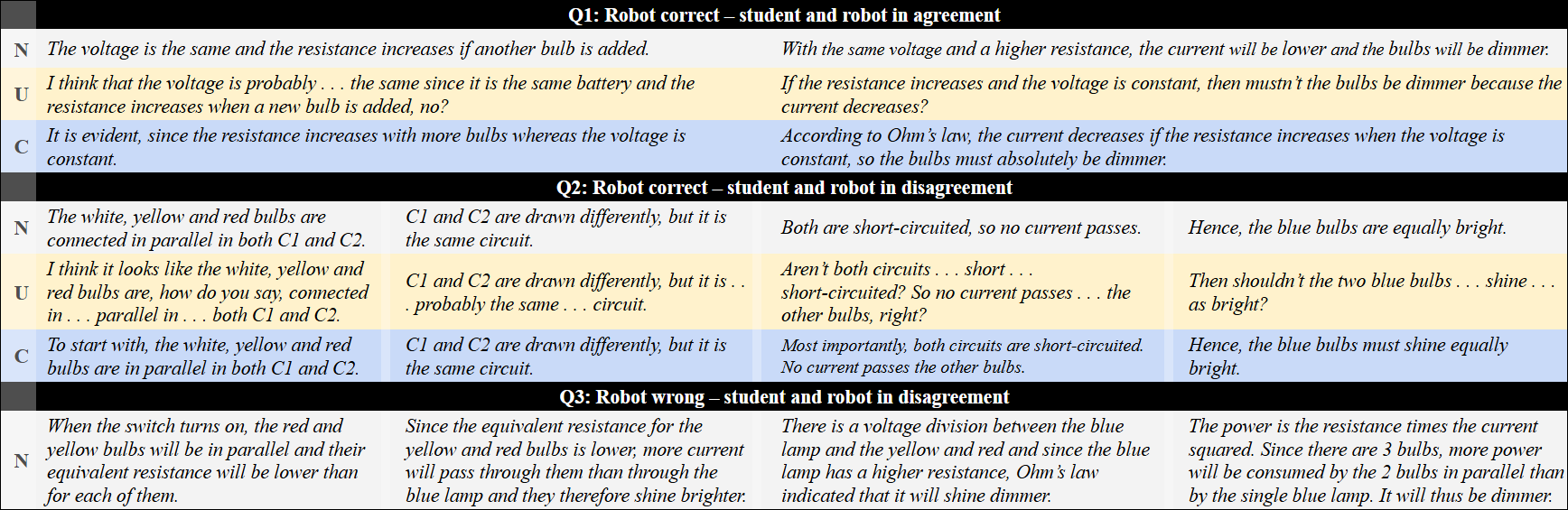}
    \end{tabular}
\end{table*}

\subsection{Electric circuit problems}
\label{sec:electronic-circuit-problems}
\noindent Based on the IRT analysis, the eight true or false problems shown in Fig.~\ref{fig:electric-problems} were selected so that the two halves of the test would have similar questions (Q1\&5, Q2\&6, Q3\&7, Q4\&8). 
The problems were four \textit{\textbf{Conditioned}} questions of \textbf{Medium} difficulty (Q1,2,5,6) with varying robot certainty levels and correct robot arguments, two \textbf{Easy} \textit{\textbf{Deception}} questions where the robot provided incorrect answers (Q3\&7) with neutral certainty, and two \textbf{Difficult} \textit{\textbf{Contrast}} questions without deception and with neutral certainty (Q4\&8).
Using the diagnostic answers (DA), we aimed for equidistant difficulty levels, but as Q4\&8 turned out to be notably more challenging for these students, the Easy (z-score: -0.82) and Medium (z:-0.43) were of more similar level than the Difficult (z: 1.67). 

\subsection{Robot Arguments for its Answer}
\label{sec:robot-arguments}
\noindent For each question, the robot first asked what preliminary answer the student had chosen (e.g., \(c1\) in Table~\ref{tab:validation-scenarios})
and agreed or disagreed with this choice (e.g., \(c2\) \& \(c3\))
and then presented a set of four arguments (e.g., \(c5\)).
The same four arguments were used for a given question, but differed in presentation, as shown in Table~\ref{tab:arguments}, depending on if the student and robot disagreed (arguments presented one by one), if they already were in agreement (arguments grouped in pairs together with expressions confirming the agreement), and the robot condition (\(U\), \(N\), \(C\)).
After presenting the four arguments, the robot asked the student to give a final answer (e.g., \(c4\) in Table~\ref{tab:validation-scenarios}).

\subsection{Pre- and post-test questionnaires}

\noindent When signing up for the experiment, the students filled in a pre-test-questionnaire (cf. the supplementary material)
at home to gather demographic data (gender, age, country of origin, preferred language of communication, educational program and grade); and answers to 10 questions based on the student characteristics questionnaire \cite{rowbotham2013} focused on the students' self-perception (using a four-point Likert scale) about their ability to learn in different circumstances; and, on five-point Likert scales, liking of STEM subjects (including Electric circuits); 8 questions from the Big Five Inventory \cite{john1991} focused on extroversion, 3 questions about trust in unknown situations, in teachers and in persons they like \cite{kaptein2009can}; 3 questions about AI attitudes (including frequency of interactions with LLMs, attitudes towards  educational robots in school and expectations regarding collaborating with a robot on school problems).

After the session, the subjects were guided to an adjacent room to fill in the post-session questionnaire 
(cf. the supplementary material)
on a tablet.
The 11 questions focused on how much confidence the students had in the robot as an exercise partner, if the student or the robot prevailed in dissent situations, on which questions dissent occurred (pictures of the electric circuits provided), how the students thought before the robot convinced them or they convinced it, the extent to which the robot influenced their thinking and made them change their mind and on which questions the robot was more and less convincing (with pictures of the circuits provided). 

\begin{figure}[b]
    \centering
    \includegraphics[width=\linewidth]
{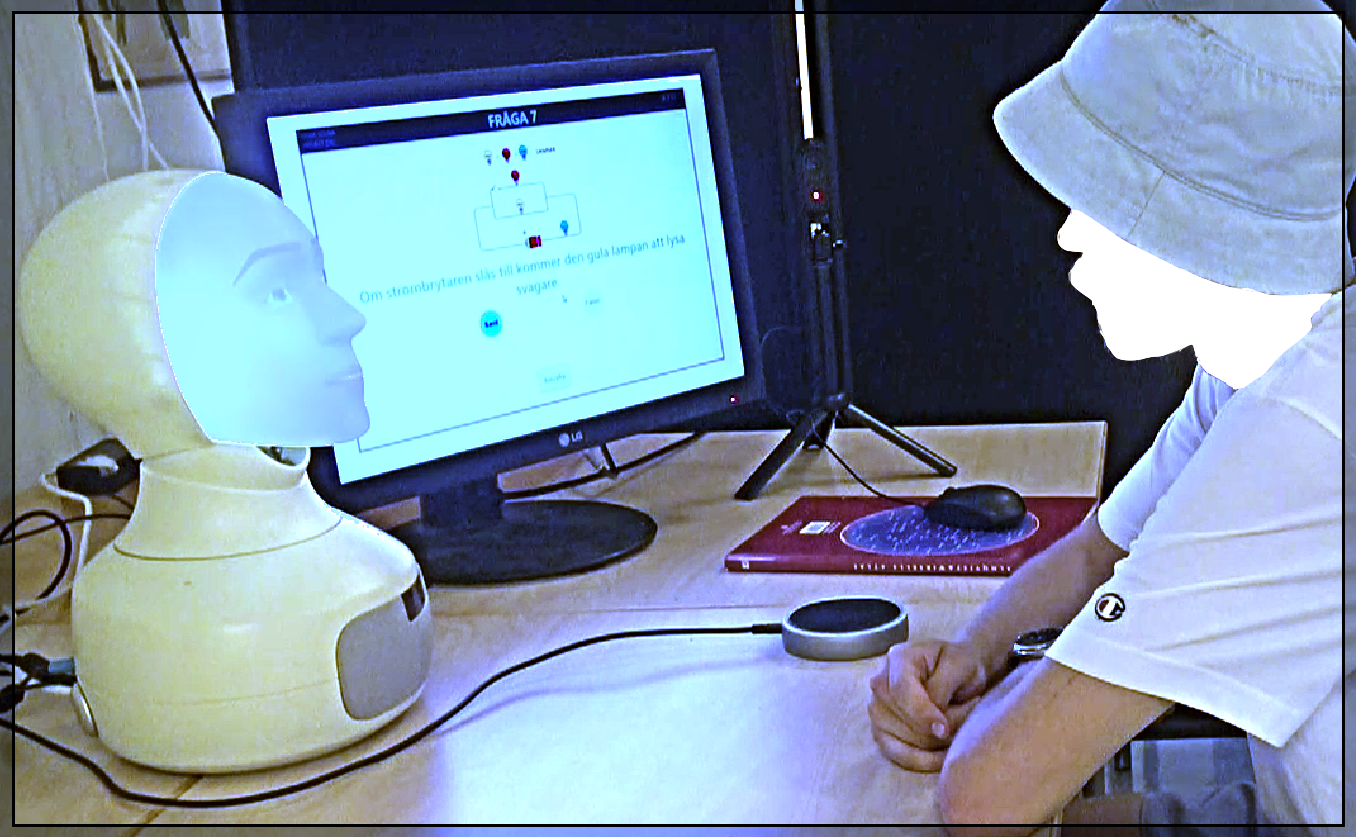}
    \caption{Experiment setup: Furhat robot (left), microphone (centre), screen with GUI and mouse (right) and tripod with camera for close-up face recording.}
    \label{fig:setup}
\end{figure}

\subsection{Procedure \& Robot Programming}
\label{sec:METHOD-setup}
\noindent The experiment was carried out in a study room at the students' school, with each student interacting individually with a female Furhat robot \cite{al2012furhat} (using the ''Isabel'' face and the Elin-Neural voice 
from Amazon Polly),  placed on a table (Fig.~\ref{fig:setup}). To the right of the robot,  a monitor presented the questions in a web-based GUI with which the students interacted using a mouse.
The system connected the robot and the GUI via a local web server using HTTP protocols, where events and commands were exchanged in JSON format. 
This allowed the robot to be contextually aware of student actions in the GUI, such as submitting a response, but to foster a natural dialogue, the robot acted as if it was unaware of the GUI interaction.

The session was recorded using a floor-standing video camera (corresponding to the view in Fig.~\ref{fig:setup}) and a table-standing cam (see Fig.~\ref{fig:setup}). 
The speech recognition results and GUI interaction data were synchronized and logged to create a multimodal interaction dataset.
One experimenter prepared the robot and the data logging for each student, gave instructions on how to initiate the interaction, and then left the room.

After a brief welcome by the robot, the GUI showed the first question and instructed to think silently before selecting TRUE or FALSE. The robot then presented its stance (Sec.~\ref{sec:robot-arguments}). After discussion,  the robot prompted students to register their final answer, with no feedback provided, before proceeding to the next question.
The robot’s responses to students' input were managed using an intent-based approach of the Furhat SDK \cite{furhatSDK}, enabling flexible handling of conversational states, such as adapting to perceived student (dis)agreement captured via speech recognition and mapped to predefined user intents. These intents were linked to corresponding states in the interaction flow.  
Video sequences of the robot's part of the interaction are available in the supplementary material.
\subsection{Statistical Analysis}
\label{sec:METHOD-ANALYSIS}
\noindent Given the likely non-normality of behavioral data, the unbalanced group sizes, and the correlated nature of repeated student observations, our primary analyses used Generalized Linear Mixed Models (GLMMs) to flexibly account for non-independence, variance heterogeneity, and to avoid assumptions of normal residuals. As covered in Sec~\ref{sec:INTRO-IRT}, IRT modeling was used to estimate students' baseline abilities and question difficulties, providing a principled way to derive expected performances without the need for traditional mixed-effects random intercepts. In peripheral analyses where data independence could be safely assumed—such as group comparisons based on single aggregate measures—we employed ANOVA models for their computational efficiency and easier interpretation. This analytical strategy allowed us to match model complexity to the nature of each analysis while ensuring that critical statistical assumptions were respected throughout.


\section{Results}\label{sec:Results}

\noindent With 40 students interacting with the robot across 8 questions, the experiment yielded 320 events, each comprising the students' preliminary answers (\textbf{PA}) and final answers (\textbf{FA}). 
The eight questions (see Fig.\ref{fig:electric-problems} and Sec.\ref{sec:electronic-circuit-problems}) included two \textit{Deception} questions (Easy), four \textit{Conditioned} (Medium), and two \textit{Contrast} (Difficult), corresponding to robot correctness and question difficulty levels as previously outlined.
The main dependent variable, \textit{Alignment, A}, is  a tri-state variable indicating PA\&FA agreement (\textit{A}=0), FA resisting the robot's influence (\textit{A}=-1) or changing to align with the robot (\textit{A}=1).


\subsection{General findings on student alignment}
\label{sec:results-general}
\noindent Students could agree with the robot either by maintaining the same PA as the robot ($n_{A=0}=181$) or by changing for their FA after hearing the robot's arguments ($n_{A\neq0}=139$).
Of the 181 instances on which the student and robot agreed on the PA, they were correct on 146 and wrong on 35.
For these 35 events, the students did not change their answer, despite hearing the robot's \textit{incorrect} arguments for this answer, which 
provided opportunities to detect errors and induce scepticism.
Dissent (\textit{A}$\neq$0) occurred 139 times: 45 for the \textit{Deception} questions (56.2\% out of 80),  55 for the \textit{Conditioned} (34.38\% of 160), and 39 for the \textit{Contrast} (48.75\% of 80).
A logistic regression model was fitted to assess the influence of question type on dissent, using \textit{Deception}~(Easy) as reference. The model revealed that dissent was significantly less likely for \textit{Conditioned}~(Medium) questions than for \textit{Deception} ($\beta = -0.898$, $p = .001$), while no significant difference was found between \textit{Contrast}~(Difficult) and \textit{Deception} ($\beta =~-0.301$, $p = .343$).

\begin{figure}
    \centering
    \includegraphics[width=0.99\columnwidth]{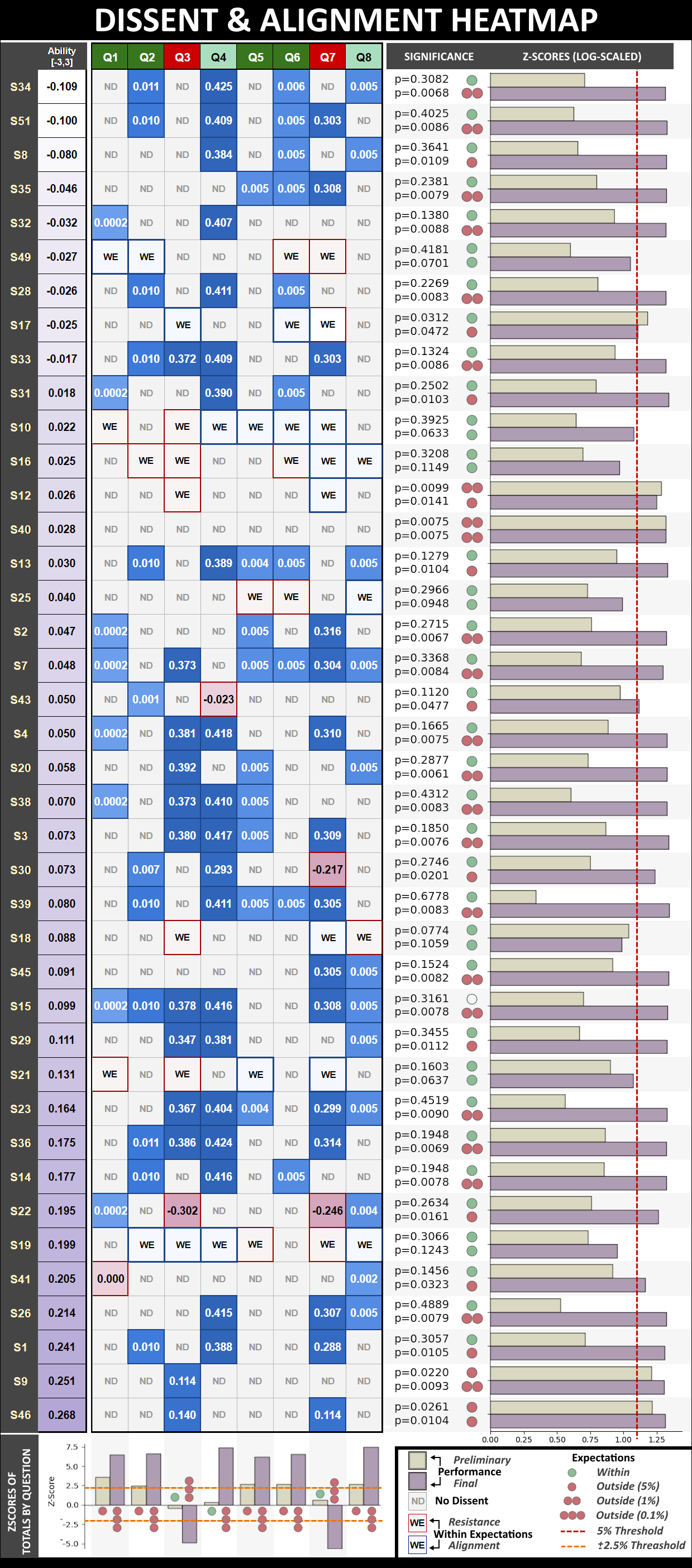}
    \caption{\textbf{(Left)} Alignment matrix between students (sorted by ability) and questions. Blue and red cells (border or filled) indicate, respectively, students aligning and resisting after PA dissent. Cell numbers describe the question-student pairs contribution to beyond expectation results. \textbf{(Right)} z-score values per student, showing deviation from expectations for PA (light bars) and FA (dark), and which students performed beyond expectations (red line and dots). \textbf{(Bottom)} z-score values per question, indicating deviation from expectations for PA and FA, and for which questions the students answered beyond expectations (red line and dots). Bars above zero indicate more correct answers than expected, below zero less, based on the diagnostic test.
    }
    \label{fig:heatmap}
\end{figure}


Students aligned with the robot after 117 of the PA dissents and resisted in only 22  (11 for \textit{Deception}, 9 for \textit{Conditioned}, and 2 for \textit{Contrast} questions), as shown in blue and red cells, respectively, in Fig.~\ref{fig:heatmap} (left). Except for one student (S40), whose PA always agreed with the robot's, all other students aligned at least once (for 25 on at least one \textit{Deception} question). 
13 students resisted the robot's persuasion at least once (10 for \textit{Deception}, with one student resisting on both Q3\&7). An analysis of variance on alignment rates found no relationship with ability (F=0.16, p=.696, R\textsuperscript{2}=0), 
showing that the robot influenced students across all ability levels.

\begin{figure}[t]
    \centering
    \includegraphics[width=1\linewidth, height=3.5cm]{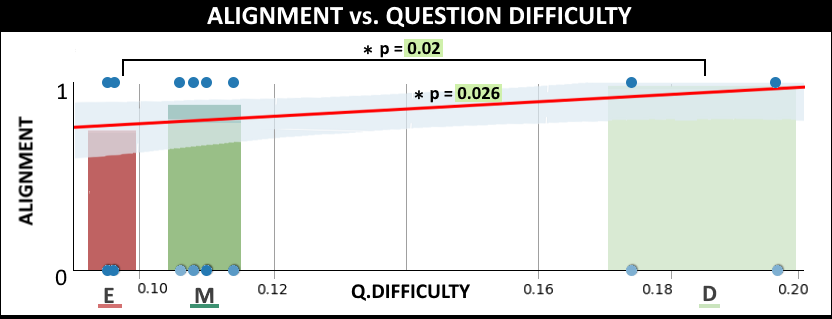}
    \caption{Student alignment as a function of question difficulty: Easy (E) \textit{Deception}, Medium (M) \textit{Conditioned}, and Difficult (D) \textit{Contrast} questions.}
    \label{fig:alignmentVsDifficulty}
\end{figure}

The linear regression analysis shown in Fig.~\ref{fig:alignmentVsDifficulty} indicated that question difficulty significantly predicts alignment ($\beta$ = 1.874, p = .\textbf{026}).
A Mixed Linear Model analysis on alignment across difficulty levels (Easy, Medium, Difficult), controlling for ability, revealed a significant increase in alignment from Easy to Difficult questions ($\beta$=-0.174, p=.\textbf{020}). There was no significant difference for Medium questions ($\beta$=-0.100, p= .161), and ability had no significant effect ($\beta$=-0.070, p=.865).
It should be noted that E questions are identical to \textit{Deception}, and thus also differ in robot correctness.


A Generalized Linear Mixed Model (\textbf{GLMM}) was used to assess the impact of \textit{Deception} \textit{vs.} non-\textit{Deception} on alignment, controlling for question difficulty, student ability, and condition as fixed covariates. The model showed no significant effect ($\beta$=-0.055, p=.469) and students did hence not respond statistically differently when the robot presented incorrect arguments. 
We implemented another model exploring interactions between these factors and \textit{Deception}, finding that question difficulty has a substantial effect size  ($\beta$=134.91, SE=101.91, p=.186) on alignment in \textit{Deception}, as will be further explored in Sec.~\ref{sec:Discussion}.



%



\subsection{Robot Influence on Student Performance} 
\noindent Following this broad overview of student alignment, we explore to what extent the changes between students’ preliminary answers (\textbf{PA}) and final answers (\textbf{FA}) were driven by the robot’s influence. Specifically, we measure how much the deviation from expected outcomes shifted between PA and FA, taking into account each student's ability.
The methodology consisted of several sequential steps:
\textit{1. Data compilation} into a unified dataset of correctness probabilities (\textit{\textbf{cp}}), PA and FA per student.
\textit{2. Expected performance per question} was calculated 
using \textit{cp} values and dispersion metrics, assessing deviations in PA and FA with p-values.
Thus, an 8-item vector per student captured IRT-derived probabilities of correctness for each question.
\textit{3. Segmented analysis} of \textit{Deception} and non-\textit{Deception} to separate negative and positive robot-induced deviations.
\textit{4. Monte Carlo simulations} estimated the probability of observed PA and FA under the null hypothesis of
random chance. Per student and question, 10,000 simulations were run based on \textit{cp} values to generate p-value distributions.
\textit{5. Fisher’s method for combined p-values} synthesized \textit{Deception} and non-\textit{Deception} questions to capture negative and positive performance shifts, enhancing statistical power \cite{Fisher1925}.
\textit{6. Deviation factors} were calculated as the difference between PA and FA p-values from steps 2 and 5, for cases where PA performance was within expectations (p$\geq$0.05) and FA performance was beyond (p$<$0.05), or where both deviated but FA performance was more extreme.




The students and questions that contributed the most to the results being beyond expectations (p$<$0.05) are indicated in Fig. \ref{fig:heatmap} (right and bottom).
An analysis revealed that 98 out of 117 alignments directly contributed to the deviation from expected results. While 35 students performed within expectations in their PA,
only seven did so in their FA. Of the 40 students, 36 deviated more in their FA than in their PA, the exceptions being S12, S17, S18, and S40.

For students S12 and S17, FA deviated less from expectations than PA, but both results were beyond expectations. 
Only student S18 performed according to expectations in both PA and FA and S40's performance remained unchanged (but beyond expectations) since there was no dissent with the robot.

\begin{figure}[t]
    \centering
    \includegraphics[width=1\linewidth,height=3.5cm]{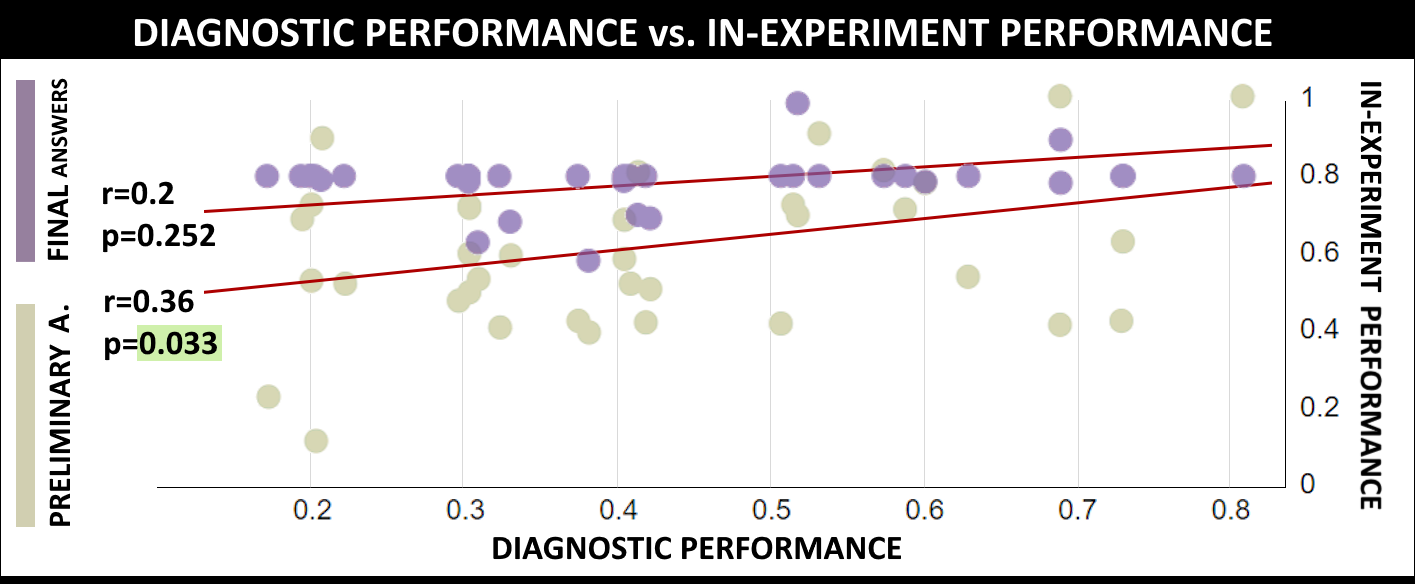}
    \caption{Comparison of Diagnostic (x-axis) and In-Experiment Performance (y-axis) on Preliminary (light dots) and Final Answers (dark dots).}
   \label{fig:DA-Pa-FA}
\end{figure}

Overall, Fig.~\ref{fig:DA-Pa-FA} shows that while there was a correlation between diagnostic and preliminary answers, establishing that the student group performed within their expected ability shown by the diagnostic test before hearing the robot's input, there was no significant correlation between diagnostic and final answers, highlighting the performance beyond expectations for FA.
A thorough and conservative investigation revealed that 75\% of the students performed beyond their expected capacity,
 \textit{above} expectations for non-\textit{Deception}  and \textit{below} expectations for \textit{Deception} questions.
This should be interpreted as a direct result of robot's influence.


\subsection{Influence of robot certainty}
\noindent We performed a series of GLMM analyses to examine the impact of the robot's displayed certainty on student alignment. We controlled for covariates student Ability and question Difficulty, and considered potential secondary effects from the sequence of robot condition. 

\begin{figure}
    \centering
    \includegraphics[width=1\linewidth, height=3.5cm]{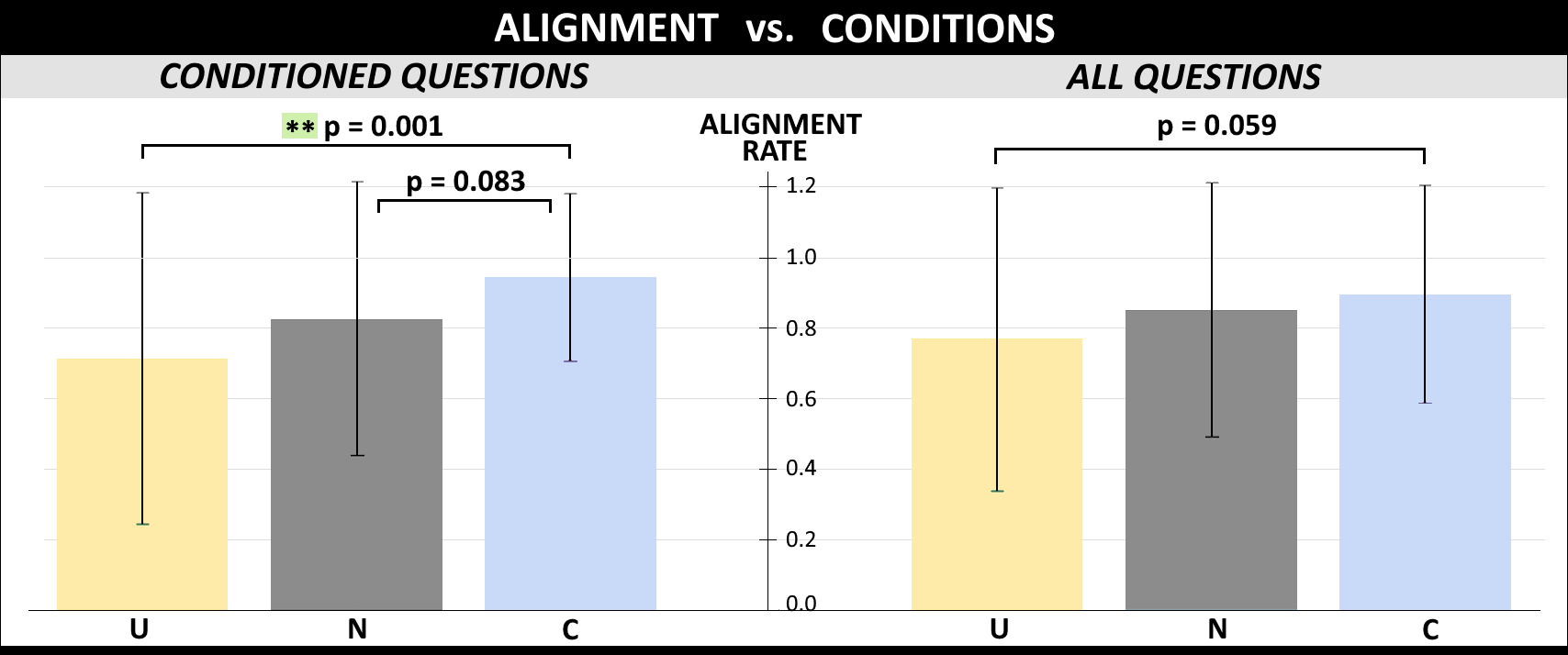}

    
    \caption{Alignment for cohorts U, N, C for (Left) \textit{Conditioned questions} and (Right) \textit{All questions}, with significance levels for differences between cohorts.}
    \label{fig:alignmentVsCondition}
\end{figure}

\textbf{\textit{Conditioned} questions:} The GLMM showed that with cohort U as reference, cohort C demonstrated significantly higher alignment (p = \textbf{0.001}), but not cohort N (p=.083), as shown in Fig.~\ref{fig:alignmentVsCondition}.
This highlights the substantial impact of the robot's expressed certainty on alignment, beyond what could be attributed to question difficulty.

\textbf{\textit{Deception} questions:} The model showed no significant effects of cohorts (\textit{U, N, C}) or ability on alignment. Coefficients for cohort and ability are close to zero (p$\gg$0.05) and group variance is minimal, indicating low variability between groups.
Since the robot portrayed neutral certainty for all cohorts in \textit{Deception}, this indicates that the robot's certainty on the two \textit{preceding} questions did not have a significant effect.

\textbf{All questions:} A Mixed Linear Model regression analysis was conducted using Alignment as dependent variable for all 139 dissents with \textit{Cohort C} as reference.
\textit{Cohort U} showed a marginal effect ($\beta$=-0.131, p=.059), suggesting a possible, although not statistically significant, reduced alignment compared to \textit{Cohort C} over \textit{all} questions (i.e., not only the ones on which the robot was displaying certainty).
The effects of \textit{Cohort N} were not significant, and as for the co-variates, alignment was significantly influenced by question difficulty ($\beta$ = 1.794, \textbf{p = .020}) but not student ability. 

\textbf{First half \textit{vs.} last:} A Mixed Linear Model regression analysis was conducted to examine the effects of robot condition, question sequence, difficulty, and ability on alignment for the 139 instances of dissent. 
The question sequence (first \textit{vs.} last four questions: $p = .907$) had no impact on alignment and interaction terms between condition and question sequence were non-significant.
This suggests that
students were not influenced by the fact that the robot had been uncertain (Group 2) or certain (Group 3) on questions in the first half when they interacted on questions in the second half.

\subsection{Pre-test questionnaire responses}
\noindent The analysis of the pre-test questionnaire targeted finding student characteristics that made them more prone to accept the robot's correct or incorrect arguments.
We first performed single-factor ANOVA on \textit{Deception} and non-\textit{Deception} questions, differentiating between alignment for students who had an above-mean response for each characteristic and those who had a below-mean.
The most striking difference is that students with more experience of using LLMs aligned 38\% more (p=\textbf{0.029}) over all questions than students with below-mean experience.
More importantly, when considering only the \textit{Deception} questions, students with more experience of LLMs were significantly more aligning with the robot's incorrect solution (see Table~\ref{tab:pre-test-analysis}).
The results in Table~\ref{tab:pre-test-analysis} further suggest that extroversion and engagement may play a role in  student alignment with the robot.
For non-\textit{Deception} questions, being full of energy and considering that efforts lead to success (which could be interpreted as being open to constructively interact with the robot on problem-solving) had a \textit{positive} impact, as these students aligned more with the robot (also supported by non-significant results for being social, +39\%, p=.065; talking a lot, +37\%, p=.075; and \textit{''Trying, I can learn anything''}, +38\%, p=.070).
On the other hand, self-perception as being positive in difficult situations and being persuasive (and thus presumably having higher confidence in one's own stance) had a marginally significant \textit{negative} effect, as this lead to, respectively, 31\% and 28\% less alignment (p=.050; p=.086).
For \textit{Deception} questions, being reserved and quiet (and presumably less prone to follow others' lead) had a \textit{positive} effect, as these students aligned less with the incorrect answer (supported by results for being shy -38\%, p=.072).

\begin{table}[b]
    \centering
      \caption{Characteristics in the pre-test responses for which there was a significant difference in alignment between students with above- and below-mean responses in the pretest for either the \textit{Deception} or non-\textit{Deception} questions.}

    \begin{tabular}{p{30mm}|ll|ll}
    \hline
    & \multicolumn{2}{c}{\textit{Deception}} & \multicolumn{2}{c}{non-\textit{Deception}}\\
    Characteristic & Conform & p & Conform & p \\
    \hline
     LLM usage   & \textbf{+71\%} & \textbf{0.038} & +26\% & 0.196\\
     Being reserved & \textbf{-49\%} & \textbf{0.016} & -1.7\% & 0.92 \\
     Being quiet & \textbf{-44\%} & \textbf{0.042} & +20\% &  0.31\\
        Efforts lead to success & +77\% & 0.092 & \textbf{+87\%} & \textbf{0.007} \\
    Being full of energy & -5\% & 0.856 & \textbf{+46\%} & \textbf{0.046}\\
        \hline
    \end{tabular}
      \label{tab:pre-test-analysis}
\end{table}

\subsection{Post-test questionnaire responses}
\noindent We focus on the students' perception of how correct and convincing the robot was, per questions and robot conditions.

Overall, the students assessed that the robot had been correct to 78\% ($\mu$=3.9/5, $\sigma$=0.87), compared to the true value of 75\%.
Groups 2 and 3, who interacted with the certain--uncertain robot, rated the robot correctness slightly (and non-significantly) higher than the control Group 1 ($\mu_1$=3.75, $\mu_2$=4.1, $\mu_3$=3.93).
The students further responded that the robot had convinced them in 84\% of the dissents ($\mu$=4.22/5, $\sigma$=0.79), compared to the true ratio of 81\% (95 out of 117 cases), with 
Group~1 perceiving that they had been convinced slightly more often ($\mu_1$=4.33, $\mu_2$=4.13, $\mu_3$=4.14).
The students' rating of the extent to which the robot made them change their mind ($\mu$=3.43, $\sigma$=1.26) and influenced their thinking ($\mu$=3.6, $\sigma$=1.27) was lower, but until a thorough ethnomethodology conversation analysis is performed, it is not possible to assess whether this discrepancy should be attributed to the students not actually being convinced by the robot when aligning, or if it is a post-test reassessment triggered by the questionnaire indicating that the robot had not always been correct.
The students were, on average, neither certain nor uncertain, before the robot convinced them ($\mu$=2.97, $\sigma$=1.32) or they convinced the robot ($\mu$=2.92, $\sigma$=1.51).

Regarding on which questions there had been a dissent and the robot had been more convincing on, two qualitative observations can be made (as the measures are the number and ratio of students mentioning each question, no statistical test is applicable).
Firstly, the ratios of question--mentions indicate that the students assessed post-test that there had been a higher degree of dissent on the \textit{Deception} ($r$=0.38) than on non-\textit{Deception} questions ($r$=0.28).
Secondly, by calculating the difference in ratio of mentions between ''more convincing'' and ''less convincing'' we find that the robot was perceived as more convincing when it was \\
    \textit{correct} ($\Delta r$=0.37) than wrong ($\Delta r$=-0.19);\\
    \textit{certain} or \textit{neutral} ($\Delta r$=0.70; 0.60) than uncertain ($\Delta r$=0.07). \\

\section{Discussion}\label{sec:Discussion}

\noindent Starting from the concept of informational trust (Sec.~\ref{sec:rw-infotrust}), we interpret the results at the event level and longitudinally.

\textbf{At the event level}, each dissent and each robot argument, constitutes an independent opportunity for alignment or resistance. A correct or incorrect robot argument allows the students to evaluate their own position and the robot’s credibility, potentially reinforcing their decision to align, strengthening their resistance or even change their previous alignment.
In 117 out of 139 instances where students' preliminary answers (PA) disagreed with the robot, they changed their final answers (FA) to align with the robot, including 34 times when it was incorrect, and 27 students \textit{always} aligned with the robot if there was a dissent, confirming \textbf{H1}. 
This high alignment rate suggests that uncertainty in their own knowledge led students to seek guidance from the robot, consistent with informational trust concepts. 
The experimental setting, the robot’s social presence, and the true/false nature of questions may have contributed to the students' uncertainty. 
Question difficulty significantly influenced alignment, with greater conformity on more difficult questions, aligning with prior research \cite{Baron1996}. 
A stringent analysis demonstrated the robot's influence, since 75\% of students performed beyond their expected capacity.


Consistent with \textbf{H2}, students were more inclined to follow the robot when it was portrayed as being certain. This effect reflects the role of source reliability \cite{cialdini2007influence}. 
When the robot expressed certainty, students likely perceived it as a knowledgeable and trustworthy source, increasing their willingness to align. Conversely, expressions of uncertainty reduced alignment, though the robot still 
exerted considerable influence, and students continued to align at a high rate.

Confirming \textbf{H4}, we found that students with more experience of using LLMs like ChatGPT were more likely to align with the robot's answers, even when the robot was incorrect. 
This indicates that prior experience with AI can enhance the perceived reliability of AI sources, potentially leading to over-reliance also in AI that are not actually driven by LLMs \cite{bender2021dangers, bommasani2021opportunities, Goddard2012}.
Personality traits may have also influenced alignment, in that students self-identifying as more outgoing and energetic were more inclined to align with the robot. 
Extrovert individuals may have a higher propensity for social engagement and be more susceptible to social influence, affecting their trust in the robot \cite{Cialdini2004}.

Alignment did not differ significantly between \textit{Deception} and non-\textit{Deception} questions. While the flawed robot arguments on Q3\&7 might have provided students with opportunities to detect inconsistencies, this was not reflected in statistically significant differences.

As these questions were the easiest on a topic that the students should master, they should have sounded incorrect or strange, prompting the students to critically evaluate and potentially resist the robot's influence. 
The rate of non-alignment in \textit{Deception} was in fact doubled compared to non-\textit{Deception} (24\% vs. 12\%), and though not statistically significant, this suggests that knowledgeable students may be more resistant when the robot's arguments are incorrect.
The rate of non-alignment in \textit{Deception} was numerically higher than in non-\textit{Deception} questions (24\% vs. 12\%), though this difference was not statistically significant. This observation may warrant further exploration in future studies examining how prior knowledge interacts with flawed arguments.

This aligns with the Elaboration Likelihood Model \cite{Petty1986}, which posits that individuals are less likely to be persuaded by weak or flawed arguments when they are motivated and able to process the information carefully. 

\textbf{Longitudinally}, Informational influence suggests individuals may adjust their perceptions based on prior experiences with the source \cite{Festinger1954,Kelley1967,Latane1981}. However, our results did not support \textbf{H3}, as no significant effects of preceding questions on student alignment were found. Statistical analyses showed that students were neither influenced by the robot's certainty on the previous two questions when confronted with the \textit{Deception} questions, nor by experiences on the first half when coming to the second. 
This suggests students treated each interaction independently, without adjusting their perceptions of the robot's reliability based on prior behaviour.

Nevertheless, two points warrant consideration. First, 
the mean change between preliminary and final answers on the \textit{Deception} questions was the largest for cohort C ($\Delta$=-0.58) and clearly smaller for cohort U ($\Delta$=-0.29), with cohort N in between ($\Delta$=-0.40), thus suggesting that it may be worth investigating carry-over effects on \textit{Deception} from preceding robot certainty in future work.
Second, we explored the behaviour of the 39 students who dissented at least once, categorizing them into \textit{Aligners} (the 32 who aligned with the robot at their first dissent) and \textit{Non-aligners} (the seven, S10, S12, S16, S18, S21, S25, S41, who did not). We calculated their alignment rate \textit{after} the first dissent and conducted an independent samples t-test. The results showed a statistically significant difference between the groups, $t(37)=3.73$, \textbf{p=.001}. Aligners ($n=32$, $\mu$=0.94, $\sigma$=0.21) had a higher Alignment rate than Non-aligners ($n$=7, $\mu$=0.58, $\sigma$=0.30).
These findings suggest that when students resist the robot at the first dissent, they may perceive the robot as less reliable and increase their self-confidence, which reduces their susceptibility to future influence \cite{Tormala2002}. 
The pattern indicates that over time, the students' interaction experiences can influence their susceptibility to informational influence.



\section{Limitations \& Future work}\label{sec:Limitations}
A number of limitations should be considered to accurately interpret the findings of this study. These can be grouped into two main areas: limitations related to the condition validation and those of the main experiment.

\textit{Robot condition validation:} 
The use of virtual robot video clips for condition validation may not fully replicate interactions with the physical robot. Additionally, the initial validation survey used decontextualized clips, potentially affecting participants' perception of the robot’s behaviour. To address these concerns, we conducted a supplementary survey with HRI experts, evaluating the certainty conveyed by the physical robot in real interaction settings.
The survey used three 80--90 second video clips presented in randomized order, each representing one certainty condition: Uncertain (\(U\)), Neutral (\(N\)), and Certain (\(C\)). The clips were extracted from experiment recordings made with the floor-standing video camera (view corresponding to Fig.~\ref{fig:setup}). Each included three key moments: the robot's answer disclosure, a claim, and a re-check of the student's stance. To focus on the robot’s behaviour, the video only showed the robot, with the student's speech muted and shortened. The videos are available as supplementary material.

We invited 47 professors and researchers with practical experience using the Furhat robot at a technical university to participate in the survey. Participants self-reported their Swedish proficiency (6-point Likert scale) and
then rated the robot’s certainty in the three clips using a 5-point Likert scale (1 = Highly Uncertain, 5 = Highly Certain). 
After data cleaning (removing 16 incomplete responses and 8 cases where videos were not fully watched), 23 valid participants remained, yielding 69 ratings per condition. Statistical analysis confirmed that \(U\) (\(\mu = 2.17, \sigma = 0.81\)) was assessed as significantly less certain than \(N\) (\(\mu = 3.91, \sigma = 0.91\)) and \(C\) (\(\mu = 4.39, \sigma = 0.67\)) (\(p < 0.001\)). The difference between \(N\) and \(C\) (\(p = 0.183\)) was not significant. 

These results support that the validation of certainty displays using the virtual robot in short video clips is valid.
The expert evaluations also relieves concerns about demographic differences between validation and experiment participants. 
Caution is nevertheless warranted when interpreting nuanced differences between conditions.


\textit{Main experiment: }
Some limitations arise from the design choices of the experiment. The study relies on Furhat’s ability to produce realistic facial expressions, which may limit reproducibility with less expressive robots. 
In the light of the finding on the influence of LLM experience, it should be noted that the present study (out of necessity to be able to control the \textit{Deception} questions) did not employ an LLM to drive the robot. A natural direction for future work would be to test how student alignment and informational trust is affected if LLMs are in fact used as the dialogue engine for an educational robot within STEM. 
Although our analysis found signs that some carry-over effects emerged during the interaction (see Sec.~\ref{sec:Discussion}), the experiment’s short time frame, set to 
avoid cognitive overload, does not fully capture longer-term interactions with the robot and longer-term effects remain unexplored.
A follow-up assessment could improve future work by determining whether the robot had a lasting impact on students’ knowledge and whether informational social influence led to internalization of the new stance  \cite{Kelman1958}.

Other limitations arose from the experiment's implementation. Since the experiment took place at the end of the academic year, motivation levels may have varied, affecting overall engagement and effort and reducing variation in ability levels. Future work should account for potential academic fatigue.  

The \textit{Deception} questions were, intentionally, easier than the others, to provoke more dissents, but since we found, non-significant, signs of a \textit{Deception}--Difficulty interaction, with a large effect size ($\beta\approx$135), future work should either vary difficulty level for \textit{Deception} question or compare them with similar non-\textit{Deception} questions to better contrast question difficulty and deception.

No-shows led to an imbalance in group sizes and, more importantly, in group ability levels, potentially affecting the comparability of conditions. Nonetheless, covariates were controlled for in all analyses.

The gender distribution of participants was uneven (8 females, 32 males), which may reduce reproducibility, as student gender could have influenced interactions with the robot \cite{widder2022gender}. Although the sample size (\(n=40\)) should be adequate for the performed analyses, increasing it to \(\geq80\) subjects could better capture variation in student ability and help mitigate the issue of gender imbalance.  

\section{Conclusions}\label{sec:Conclusion}
\noindent Referring back to the title, we found that 30 out of 40 students showed little \textit{sense} in assessing the robot's arguments for the easy \textit{Deception} questions, aligning with the robot's wrong answer on both question, and 9 other students aligned on one of them. Only one student resisted on both.
Further, more experience of LLMs reduced students' critical assessment of robot arguments.
On the other hand, the robot also influenced the students to perform significantly better than their ability on non-\textit{Deception} questions.
The students did demonstrate  \textit{sensibility} in responding to the robot's displays of certainty, with significantly higher alignment with the robot's -- correct -- answer when it was portrayed as being certain than when it was portrayed as uncertain.

The implications of this study are that educators must be aware of the high informational trust that students have in general in educational AI and that influence of AI experience increases this trust.
Developers must also enable social educational robots with means to signal certainty or uncertainty in presented facts.
As LLMs begin to provide reliability metrics \cite{manakul-etal-2023-selfcheckgpt}, robots should make use of both the voice \cite{Miniota2023} and familiar human facial expressions \cite{Vincze2016} to express less certainty for potentially unreliable information to reduce student susceptibility to misinformation and enhance critical thinking.


\section{Ethical approval}\label{sec:Ethical approval}
\noindent The research has undergone ethical review by the Swedish Ethical Review Authority 

\section{Funding}
This work is supported by the Marcus and Amalia Wallenberg Foundation under grant MAW 2020.0052 and the Swedish Research Council (VR) under grant 2022-03265.


\bibliographystyle{plainnat}
\bibliography{sn-bibliography}


\end{document}